\documentclass[11pt,a4paper]{article}
\usepackage[hyperref]{acl2020}
\usepackage{times}
\usepackage{latexsym}
\usepackage{graphicx}
\usepackage{comment}

\usepackage{microtype}

\aclfinalcopy 

\title{How are you? Introducing stress-based text tailoring}

\author{Simone Balloccu, Ehud Reiter \\
  University Of Aberdeen \\ United Kingdom \\
  \texttt{simone.balloccu@abdn.ac.uk}\\
  \texttt{e.reiter@abdn.ac.uk} \And 
  Alexandra Johnstone, Claire Fyfe \\
  Rowett Institute, University of Aberdeen \\ United Kingdom \\
  \texttt{alex.johnstone@abdn.ac.uk}\\
  \texttt{c.fyfe@abdn.ac.uk}
  }
\date{}

\begin{document}
\maketitle
\begin{abstract}
Can stress affect not only your life but also how you read and interpret a text? Healthcare has shown evidence of such dynamics and in this short paper we discuss customising texts based on user stress level, as it could represent a critical factor when it comes to user engagement and behavioural change. We first show a real-world example in which user behaviour is influenced by stress, then, after discussing which tools can be employed to assess and measure it, we propose an initial method for tailoring the document by exploiting complexity reduction and affect enforcement. The result is a short and encouraging text which requires less commitment to be read and understood. We believe this work in progress can raise some interesting questions on a topic that is often overlooked in NLG.
\end{abstract}

\section{Introduction}
Healthcare can benefit greatly from NLG as communication plays a big role in therapy outcome \cite{stewart1995effective} and electronic documentation is useful \cite{ross2016factors} but heavy and time-consuming to be produced \cite{arndt2017tethered}. NLG can speed up the process and there is evidence of increasing adoption of data-to-text in healthcare \cite{pauws2019making}. Despite NLG being valid for producing standardised documentation, the situation is more complex when the targeted reader is a patient: people are all different and a single text is often not enough. Healthcare-NLG often addresses the problem by adopting user tailoring, which is the practice of generating a text customised to the user traits and preferences \cite{Kukafka2005}. In this work, we focus on user stress, a factor which is often not considered in NLG, as a feature for text-tailoring. We show that stress does matter when reading text and provide a real-world example (the dietary domain) along with some initial discussion and examples on how stress-based tailoring could be carried out. Motivated by the PhilHumans project\footnote{PhilHumans Project page: https://www.philhumans.eu/} and its goal of improving the interactions between personal health apps and users, we explore the benefits of personalising generated texts based on how much stressed the user is. Stress impacts people’s health, behaviour, receptivity to advice, and the ability to absorb complex information, so basing tailoring on this element could help health software support patients better. The structure of the paper is the following: in Section \ref{sota} we recap the state of user tailoring in NLG, emphasizing that no known system does properly address user stress; in Section \ref{neurofast}-\ref{analysis} we provide a real-world example in the dietary domain, showing the results of statistical analysis that, in line with previous healthcare studies, hints at the correlation between stress and eating behaviour; Section \ref{stressassessment} discusses possible methodologies to assess user stress, prioritising the non-intrusive ones; in Section \ref{tailoring} we propose two theoretical ways to tailor the text for a stressed user, along with an example; Section \ref{conclusion} closes the paper with our final considerations.

\section{Related Work}
\label{sota}
Natural Language Generation has been used in the healthcare domain in a variety of ways, with different aims. Since the spread of e-health, various works have focused on automating the creation of documents. some examples of these are clinical encounters \cite{finley2018automated}, chief complaints \cite{lee2018natural} and handover reports \cite{schneider2013mime}. There has also been some research on producing multiple texts from the same source, as in the BabyTalk project \cite{portet2008babytalk} which generated premature infants reports for doctors\cite{portet2009automatic}, nurses \cite{hunter2011bt} and parents \cite{mahamood2011generating} in Neonatal Intensive Care Unit (NICU) environment. Another scenario in which NLG contributed to healthcare is decision support \cite{binsted1995generating} \cite{hommes2019personalized}, where the text can help patients understand their clinical situation, and prepare them for shared decision making \cite{seminar2011salzburg} \cite{stiggelbout2012shared}. The majority of these systems don't need any personal information about the reader because their final purpose is purely informative. It is, however, different when we move to behavioural change, as the text aims to trigger a lifestyle improvement in the reader. In this case we're dealing with persuasive communication, which is the act of communicating to make the reader perform certain
actions or collaborate in various activities \cite{guerini2011approaches}. Persuasive communication systems can be distinguished in vertical and horizontal approaches \cite{maimone2018perkapp}: vertical systems tends to be specialised, hence they are very effective but harder to move between different domains, while horizontal systems are often conceptual and focused on domain-independent knowledge and strategies. When persuasive communication has to be obtained through NLG, a popular approach is to profile the user to model the text accordingly. This practice is known as text tailoring \cite{Kukafka2005} and has shown promising results \cite{kreuter2003tailored}. Morevoer NLG tailoring is particularly helpful, since manual text customisation is often time-consuming \cite{pauws2019making}. NLG-driven tailoring was tested for lifestyle improvements and behavioural change in various domains. Vertical approaches includes solutions for smoking cessation \cite{reiter2003lessons}, driving conduct improvement \cite{braun2018saferdrive}, diet management \cite{anselma2018checkyourmeal}\cite{anselma2018designing}\cite{anselma2017approach} and therapy recommendation \cite{skinner1994physicians}. Horizontal approaches deals with more general topics, like the key challenges and model design for behaviour change \cite{oinas2013foundation}\cite{kelders2016health}\cite{oinas2009persuasive}, argumentative persuasive communication \cite{zukerman2000using} \cite{reed1996architecture} or the inclusion of generic psychological traits and basic human values into the model \cite{ding2016personalized}. Finally, there have been some attempts to partially merge horizontal and vertical models \cite{maimone2018perkapp}\cite{dragoni2018horus}\cite{donadello2019persuasive}. When the text must be built according to user traits, it is important to identify the variables which could influence not only the behaviour but also user engagement. These include general psychological tailoring techiques \cite{hawkins2008understanding}; user numeracy, literacy \cite{williams2008generating} and domain knowledge \cite{mckeown1993tailoring}\cite{paris1988tailoring}; affect \cite{mahamood2011generating}. User stress is a factor which, to our knowledge, is largely ignored in the existing literature. Some NLG works have considered stress, but not in a dynamic way. Some works in tactical NLG (text which tries to induce a certain emotional status) \cite{van2008towards} considered stress, but only to assess the text effect. This makes sense since tactical NLG aims to induce an emotional status instead of detecting or managing it. BabyTalk-Family project \cite{mahamood2011generating} tried to actively tailor the document by estimating stress value, but without detecting it from the user itself. Again, the choice is reasonable since having children in NICU can be a traumatic experience, therefore directly asking questions about stress could be intrusive and dangerous. Beside these boundary cases we argue that stress-based tailoring is of primary importance as stress has been shown to significantly impact individual reading skills \cite{rai2015effects}\cite{peng2018meta}, potentially causing temporary difficulty in understanding the given text. Therefore stress data can be exploited to guide the tailoring process with precious hints on how to re-realise the text. This motivates researching how user stress assessment can be done in the least intrusive way.

\section{A real world example: NeuroFAST}
\label{neurofast}
Before going into the details of how stress-based tailoring could be achieved, we want to introduce some preliminary evidence that stress does matter and can influence user behaviour. We exploit a dataset which is derived from the EU-funded project NeuroFAST\footnote{NeuroFAST project page: https://cordis.europa.eu/project/id/245009}, regarding the socio-psychological forces that could influence eating behaviour. Our dataset \cite{malone2015effects} contains a detailed food diary for 413 different workers, together with a ``Daily Hassles" questionnaire in which the participants noted stressful events. Each entry is scored in a range between 0 and 4, with a higher score indicating a higher stress level. For example, we can find an individual that put ``Caught in a traffic jam on the way to work” as a daily hassle. This resulted in a score of 2/4. We analysed the dataset to test the following hypothesis: does stress influence workers' diet? This is not a novelty in healthcare, as different studies linked stress to eating disorders \cite{scott2012stress}\cite{torres2007relationship}\cite{puddephatt2020eating}\cite{riffer2019relationship} but we still provide an additional, post hoc, analysis on said data to show a concrete example. Additionally, even if the original study was not meant to inspect such correlation, both data were readily available, making it possible to inspect it.

\subsection{NeuroFAST stress analysis}
\label{analysis}
To test our hypotheses we extracted the data, isolating the absolute calorie change for every couple of days: this was necessary since the dataset itself doesn't provide any kind of dietary goal (such as the daily calorie target) and hence we could only calculate how each participant energy intake changed during the week. This gave use 1410 data points, each one represented as a pair in which the first element is the absolute energy shift (how much did the intake increase/decrease compared to the previous day?) and the second one is the daily stress score. Overall, the dataset contained only one week worth of data for each worker, so we couldn't further inspect their eating behaviour. We also had to cut out some participants cause they didn't fill in the ``Daily Hassles" form. Now we proceed to show the results of our statistical analysis, to test whether stress does affect individual diet or not. It must be considered that a higher stress score doesn't necessarily imply overeating: people react differently to stress (some will eat more, other will eat less) and that's another reason why we calculated the absolute calorie change. For the first analysis, we looked for a correlation between the calorie shift and some stress trends which can occur in workers' life. We identified four stress patterns by comparing the score for every given pair of days:

\begin{itemize}
    \item \textbf{Stress variation:} the score changes (increases or decreases) while staying positive
    \item \textbf{Stress drop:} the score drops to zero from any given value
    \item \textbf{Stress stasis:} the score doesn't change but stays positive
    \item \textbf{Stress absence:} the score stays equals to zero during the two days
\end{itemize}

By defining these patterns we were able to check if a particular stress trend could be related to a calorie shift. We ran a one-way ANOVA which showed statistically significant difference in the distributions (p $\approx$ \textbf{0.008}; F-value  = \textbf{3.9}; Df = \textbf{3}; $\eta^{2} \approx$ \textbf{0.008}). An additional Tukey Honestly Significant Difference (HSD) test revealed that the only statistically significant difference was between the Stress Absence and Stress Stasis trends (\textbf{p $\approx$ 0.031}). In other words, there was a significant difference in how people ate when they were consistently stressed during the weeks compared to when they weren't stressed. The analysis results are summarized in Table \ref{table1}.

\begin{table}
\begin{center}
{\caption{Stress trend analysis result (ONE-WAY ANOVA)}\label{table1}}
\begin{tabular}{|c|c|}
\hline
\textbf{Trend}&\textbf{p-value}
\\
\hline
\\[-8pt]
\quad variation - absence & 0.520\\
\hline
\quad drop - absence & 0.886\\
\hline
\quad \textbf{stasis - absence} & \textbf{0.0317}\\
\hline
\quad drop - variation & 0.326\\
\hline
\quad stasis - variation & 0.722\\
\hline
\quad stasis - drop & 0.052\\
\hline
\end{tabular}
\end{center}
\end{table}

For the second analysis we considered a simpler scenario: we aggregated every stress trend except for the absence, to inspect a link between being generally stressed in any possible way or calm and the calorie shift. In this regards we ran a Welch Two-Sample T-Test which gave us a significant result (p $\approx$ \textbf{0.046}). We want to remind that the analysis we ran is a post-hoc one: original experiment wasn't mean to test our hypotheses. Hence our results can only be taken as an hint that stress could have influenced participants' eating habits. 

\section{Stress-based tailoring}
\begin{figure}
\includegraphics[scale=0.7]{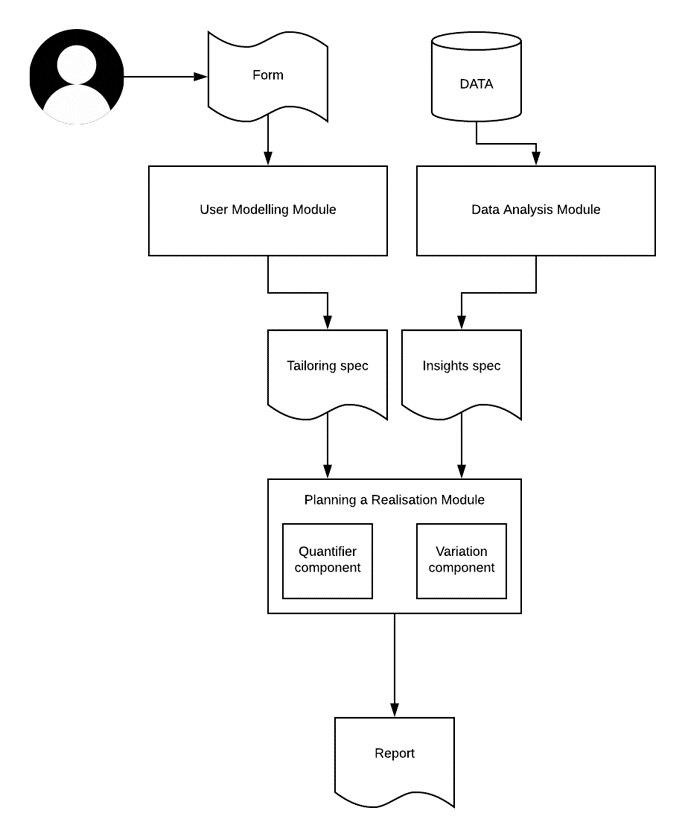}
\caption{Original framework architecture} \label{architecture}
\end{figure}

We now proceed to detail a hypothetical way to tailor a text-based on user stress. To do so we do stay in the dietary domain and consider producing a diet-coaching report by using NLG. To do so, we take into consideration our previously proposed architecture for dynamic tailoring in healthcare \cite{balloccu2020nlg}, which we summarise in Figure \ref{architecture}. The framework is designed to extract data regarding user diet, then produce a text which is tailored by taking into account both general tailoring techniques and user individual traits. What we do in here is to formulate a theoretical way to enrich the tailoring logic with details about the user stress which will guide the size and content of the final text.

\subsection{Assessing user's stress}
\label{stressassessment}
 As a first step, we must determine whether the user is stressed and find a measure for this data: this is a major challenge, as it should be done in the most reliable and least intrusive possible way. Detecting stress is usually pursued through two different tools: body sensors and self-assessment tools\cite{plarre2011continuous}. Intrusion-wise, sensors are the least desirable approach: since our target is to deliver a report with a high frequency, it is unrealistic to expect the user to wear detection devices daily. Even if the high-frequency constraint is omitted, sensors tend to be reliable in controlled environments only \cite{healey2005detecting}\cite{madan2011sensing}\cite{plarre2011continuous}, as they tend to struggle to distinguish stressful events from normal activities which alter user's psychophysical condition (like sports activities). On the other hand, self-assessment tools typically involve questionnaires like DASS-21 \cite{lovibond1995structure} or PANAS\cite{watson1988development}. Such tools can produce an estimation of users' stress related to a relatively short period (a week for example). Moreover, it could be even possible to assess daily stress: this would make the system capable of tailoring the text ``on the fly", potentially fine-tuning the text details every single day. A possible way to address daily stress is, for example, the ``Daily Hassles" form that was used in NeuroFAST project. However, measuring stress every day reduces self-assessment tools reliability, and reduces their validity to some specific scenarios \cite{adams2014towards} only. Finally, it must be considered that forcing the user to fill a form every single day could end up being a stressful operation as well. From a discussion with experts from NeuroFAST project, it turned out that the stress values were collected by using a mix of sensors and forms, an experience which the medical staff itself described as potentially stressful for the participants, especially when they were shift workers.

\subsection{Enriching user tailoring with stress}
\label{tailoring}
Given what has been said until now, we proceed to formulate two key parameters on top of which the tailoring will take place:
\hfill \\
\begin{itemize}
    \item \textbf{Complexity:} this involves both the length and terminology of the given text. There have been proofs that stress does impact on working memory \cite{rai2015effects}\cite{peng2018meta}. This means that being stressed potentially influences reading skills. Addressing this scenario can be done by delivering a shorter text and simplifying complex terminology, in line with previous work \cite{mckeown1993tailoring}\cite{paris1988tailoring}.
    \item \textbf{Affect:} when dealing with stressful events, like work or personal problems, \cite{scott2012stress}\cite{torres2007relationship}\cite{puddephatt2020eating}\cite{riffer2019relationship} users could feel demoralised after reading a report which states they didn't meet their current goals. This applies well to the dietary domain but it might reasonably make sense in other areas. A possible way to address this problem is adopting an affective tone in the text, in line with previous studies \cite{mahamood2011generating}\cite{de1999affective}. Omitting certain information could be a valid option as well under some circumstances. For example \cite{van2018lying}, showed how "lying" could be useful in the NICU environment, since preventing the child grandparents from knowing about negative development could have avoided potential health issues (i.e. heart attacks).
\end{itemize}
\hfill \\
Regarding complexity, good tailoring should address both reducing text length and simplifying the content itself. This can be split into two complementary document planning (length) and micro-planning (terminology) steps. This is especially important given that, even without considering stress, patients typically struggle to understand complex medical language \cite{elhadad2006comprehending}\cite{zeng2008estimating}. We leave terminology simplification as future work and focus, for now, on obtaining a shorter text. We do consider an output example from our system (Figure \ref{output}), which informs the user with some weekly insights about his/her diet. The text itself is lengthy, hence we try to shorten it to get an immediately understandable content. To do so we remove superfluous information which doesn't contribute to giving the user facts about their diet. A hypothetical result is showed in Figure \ref{reduced}. The text is now around a quarter of its original length but delivers the same information (i.e.: calories and nutrients). Potentially traumatic information was removed, like adverse effects, and some of the tailoring techniques got lost during the shortening phase: these include both psychological practices \cite{hawkins2008understanding} and user choices like adverse effects. We argue that, by sacrificing a bit of customisation, we're gaining a lot of readability which is a critical parameter when the user is stressed. Moreover, the majority of user choices are still being kept. It is, however, fair to underline that the system should offer the user the chance to see the original uncut text, to access the omitted details. It is now necessary to re-introduce the affective bits inside the text as stated before: despite the new text being much faster to be read in terms of length, it sounds quite cold compared to the original one. Therefore we re-insert some motivational frames without being redundant and with the primary target of keeping the text short and simple. A mockup with an affect component can be seen in Figure \ref{affect}: the improved strategy is pretty much the same we originally adopted \cite{balloccu2020nlg} and focuses on lowering the weight of bad news and encouraging the user to pursue their goals (which were re-introduced into the text). Overall the text is still significantly shorter than the original one.

\begin{figure}[h]
    \fbox{\begin{minipage}[f]{20em}
    Hi Paul, this is your weekly diet report. You told us that you want to lose weight and gain self-confidence. So we wrote this to help you out.\\
    ~\\
    You did a great job on calories; you had some problems with sugar and sodium.\\
    ~\\
    Going a bit deeper, Monday was your best day! Your calories were about perfect then. Friday gave you some problems, as you ate about a third more than what you should.\\
    ~\\
    Your sugar consumption is around twice more than what it should be, and much of it came from Coca Cola. We know it's hard to change what you eat, but sugar excess is bad for your health and can cause pale skin, anxiety and fatigue. Try having less sugary foods as it would turn in weight loss, less dental plaque and more energy.\\ 
    ~\\
    Your sodium consumption is around half more than what it should be. A lot of it came from Pringles chips. Keep in mind that sodium excess can lead to nausea, headache and seizures. Less salty foods mean memory improvement, less bloating and lower blood pressure.\\
    ~\\
    Paul, we came up with some suggestions based on your needs. Regarding sugar, you could replace sugary drinks with tea or coffee. Also, sodium will be lower if you avoid salty snacks and opt for fruits instead.
    \end{minipage}}
    \caption{\label{output}Starting text}

    \fbox{\begin{minipage}[f]{20em}
    Hi Paul, calories were good this week, especially Monday, but you had problems on Friday. Nutrient-wise, you had around twice more than your sugar target (cut a bit on Coca-Cola) and around a half more sodium than what you should get (try reducing chips).\\
    \vspace{-3mm}
    \end{minipage}}
    \caption{\label{reduced}Reduced example}

    \fbox{\begin{minipage}[f]{20em}
    Hi Paul, this week you did a great job on calories, especially Monday. You had some problem Friday: we're sure that next week will be better! Nutrient-wise, you had around twice more than your sugar target (cut a bit on Coca-Cola) and around a half more sodium than what you should get (try reducing chips). Changing what you eat is a long path, keep up will succeed in losing weight and gaining self-confidence!\\
    \vspace{-3mm}
    \end{minipage}}
    \caption{\label{affect}Reduced example + affect}
\end{figure}
\clearpage

\section{Conclusion and future developments}
\label{conclusion}
In this short paper, we highlighted the importance of addressing user stress when performing text tailoring. To the best of our knowledge there's not a single system which tries to use stress as a variable to tailor textual generation. We think that ignoring this factor prevents current systems from reaching a further step towards user engagement, especially in the domain of behavioural change, as stress does impact both reading capabilities and user motivation. We showed a real-world case in which stress does matter, the NeuroFAST project: our post hoc statistical analysis hints at the fact that the workers who took part into the experiment changed their diet significantly when stressed. This result is aligned with previous well-known evidence which state that diet and stress are linked. We then took our previously proposed framework, which dynamically tailors the user to communicate health data, and enriching its components to exploit stress value as an additional tailoring parameter. By doing so, we showed a few preliminary ideas on tailoring text based on the stress level. The new tailoring logic is mainly achieved by reducing the text size, hence drastically reducing the information delivery time, and enforcing its affective component to counterbalance the lack of motivation which stress can cause when following a diet. 

\subsection{Implementation}
Given that this is a work-in-progress project and all of the given examples are mockup (no computational steps or formal definition of the problem were given, as this paper is purely conceptual), we intend to proceed to work on this thematic, including the implementation of a proper stress-based tailoring system to evaluate its effectiveness. Moreover, the proposed text lacked terminology simplification, a phase which is particularly important in healthcare, where the patient often ignores the meaning of medical terminology. Along with an actual system implementation, it is in our interest to design an evaluation framework, which could help us get a realistic idea of the tailoring strategy effects. This would also include considering and inspecting all of the ethical challenges that comes when working with stressed people.

\subsection{Exploring different domains}
We focused on the dietary domain, but we remind that behaviour change is a vast domain and we're looking forward applying stress-based tailoring in different areas: at the moment we believe that sleeping apnea therapy (CPAP) could take great advantage of this technique as sleep deprivation is indeed a stressful event and tailored text showed to be a promising tool in increasing therapy adherence\cite{tatousek2016tailor}. Overall we do hope that this work raised some interesting research questions around a variable which is, as for now, not considered in NLG and user-tailoring. 

\subsection{Acknowledgments}
This research was funded by the European Union’s Horizon 2020 research and innovation programme under the Marie Skłodowska-Curie grant agreement No 812882.
\bigskip
\\The research leading to these results has received funding from the European Community’s Seventh Framework Programme FP7-KBBE-2010-4 under grant agreement No: 266408. Collaborators from the University of Aberdeen, Rowett Institute gratefully acknowledge financial support from the Scottish Government as part of the Strategic Research Programme at the Rowett Institute.

\bibliography{bibliography}
\bibliographystyle{acl_natbib}

\end{document}